\documentclass{article}
\pdfoutput=1
\usepackage[preprint]{spconf}
\usepackage{amsmath,graphicx}
\usepackage{subfigure}
\usepackage{amsfonts}
\usepackage[ruled]{algorithm2e}
\usepackage{algorithmic,algorithm2e}
\usepackage{booktabs}
\usepackage{multirow}
\usepackage{color}
\usepackage{floatrow}
\usepackage{comment}
\usepackage{marvosym}
\usepackage{caption}
\usepackage{xcolor}

\def\x{{\mathbf x}}

\title{Hypergraph-MLP: Learning on Hypergraphs without Message Passing}
\name{Bohan Tang\textsuperscript{1}, Siheng Chen\textsuperscript{2,3}, Xiaowen Dong\textsuperscript{1}}
\address{\textsuperscript{1}University of Oxford, \textsuperscript{2}Shanghai Jiao Tong University, \textsuperscript{3}Shanghai AI Laboratory}
\ninept 

\newcommand{\R}{\ensuremath{\mathbb{R}}}

\def\Z{\mathbf{Z}}
\def\X{\mathbf{X}}
\def\Y{\mathbf{Y}}
\def\x{\mathbf{x}}
\def\z{\mathbf{z}}
\def\W{\mathbf{W}}

\def\y{\mathbf{y}}
\def\hH{\mathbf{H}}

\def\V{\mathcal{V}}
\def\E{\mathcal{E}}
\def\hhH{\mathcal{H}}
\def\G{\mathcal{G}}

\def\N{\mathcal{N}}

\def\LL{\mathbf{L}}
\def\ttt{\mathbf{\Theta}}

\DeclareMathOperator{\diag}{diag}

\begin{document}
\maketitle
%\tableofcontents
%
\begin{abstract}
Hypergraphs are vital in modelling data with higher-order relations containing more than two entities, gaining prominence in machine learning and signal processing. Many hypergraph neural networks leverage message passing over hypergraph structures to enhance node representation learning, yielding impressive performances in tasks like hypergraph node classification. However, these message-passing-based models face several challenges, including oversmoothing as well as high latency and sensitivity to structural perturbations at inference time. To tackle those challenges, we propose an alternative approach where we integrate the information about hypergraph structures into training supervision without explicit message passing, thus also removing the reliance on it at inference. Specifically, we introduce Hypergraph-MLP, a novel learning framework for hypergraph-structured data, where the learning model is a straightforward multilayer perceptron (MLP) supervised by a loss function based on a notion of signal smoothness on hypergraphs. Experiments on hypergraph node classification tasks demonstrate that Hypergraph-MLP achieves competitive performance compared to existing baselines, and is considerably faster and more robust against structural perturbations at inference.
\end{abstract}
\section{Introduction}
\label{sec:intro}
Hypergraphs, comprising nodes and hyperedges, offer a versatile extension of graphs, effectively capturing higher-order relations among multiple nodes~\cite{bick2023higher,xu2022GroupNet,tang2023learning,xu2023dynamic}. Recently, efficiently processing hypergraph-structured data has garnered significant attention in machine learning and signal processing~\cite{zhang2019introducing,antelmisurvey}. In the literature, hypergraph neural networks, leveraging message passing over the inherent hypergraph structure to facilitate effective node representation learning, excel in various tasks, e.g., hypergraph node classification. Nonetheless, these models exhibit a series of limitations, akin to those observed in classical message passing neural networks (MPNNs). These include oversmoothing~\cite{chen2022preventing, rusch2023survey}, high latency~\cite{antelmisurvey,zhang2022graphless}, and sensitivity to structural perturbations~\cite{sun2022adversarial, hu2023hyperattack} at inference time.

\begin{figure}[t] 
\centering
\includegraphics[width=.78\textwidth]{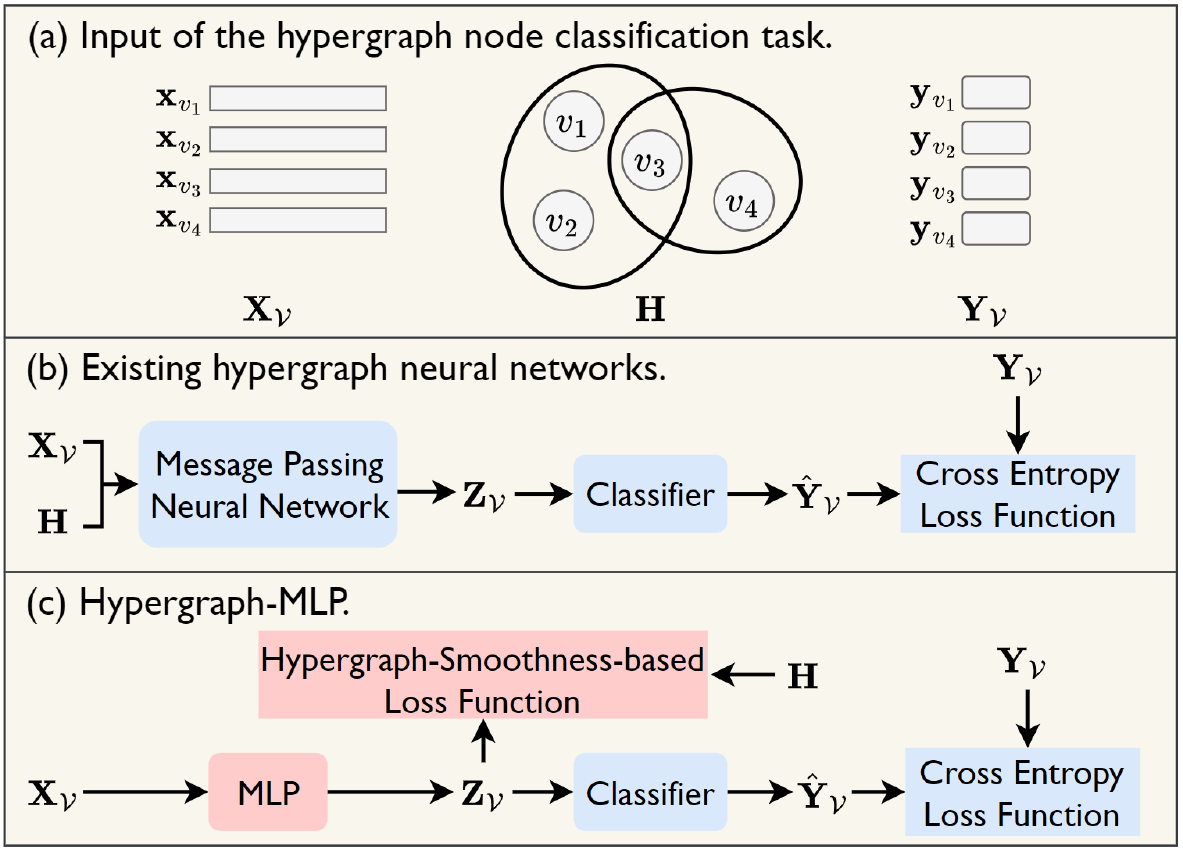}
\vskip -0.01in
\caption{The training paradigm of the Hypergraph-MLP and hypergraph neural networks for hypergraph node classification, where $\X_{\V}$ denotes node features, $\hH$ is the incidence matrix, $\Y_{\V}$ denotes node labels, and only partial labels in $\Y_{\V}$ are used at training.} %Key components of the Hypergraph-MLP are highlighted in red.} 
\label{fig:compare_hgnn}
\vskip -0.1in
\end{figure}

To address these limitations, our key idea is to integrate the hypergraph structure information into training supervision, rather than using it for message
passing. This leads to a novel learning framework for hypergraph-structured data, which we call Hypergraph-MLP. Our framework consists of two key components: an MLP-based model devoid of message-passing operations and a hypergraph-smoothness-based loss function rooted in the smoothness prior, inspired by~\cite{tang2023hypergraph}, assuming that embeddings of nodes in a hyperedge are highly correlated via the hyperedge connecting these nodes. Training with the hypergraph-smoothness-based loss function allows the MLP-based model to generate node embeddings aligned with the aforesaid smoothness prior, effectively using structural information without doing message passing. The advantages of Hypergraph-MLP over existing message-passing-based hypergraph neural networks are threefold. First, without message passing operations, the Hypergraph-MLP inherently avoids the oversmoothing issue. Second, Hypergraph-MLP uses feed-forward propagation with a complexity of $\mathcal{O}(Ln)$, but message-passing-based models have a complexity of $\mathcal{O}(Ln + Lm)$, with $n$ as the node count, $m$ as the hyperedge count, and $L$ as the layer count. This leads to lower inference latency in Hypergraph-MLP compared to those models. Third, removing the reliance on the hypergraph structure at inference makes Hypergraph-MLP more robust to structural perturbations. As an overview, Fig.~\ref{fig:compare_hgnn} compares the architectures of the Hypergraph-MLP and existing hypergraph neural networks.

Our contributions are twofold. First, we propose Hypergraph-MLP, the first learning framework for hypergraph-structured data that efficiently incorporates the hypergraph structural information into supervision instead of using the structure for message passing. It provides a new paradigm for designing neural networks to process data on hypergraphs. Second, we extensively validate Hypergraph-MLP in the hypergraph node classification task. The results show that, compared to existing hypergraph neural networks, Hypergraph-MLP achieves competitive accuracy, fastest inference, and better robustness against structural perturbations at inference.

\section{Related Works}
\textbf{Learning on hypergraphs.} There are many useful tools for modelling data with higher-order interactions, e.g., hypergraphs, and simplicial complex~\cite{bodnar2021weisfeiler,battiloro2023generalized,hajijtopological}. This paper focuses on the process of hypergraph-structured data. Recent research~\cite{chien2022you} shows that most neural networks designed for hypergraphs follow a two-stage message passing paradigm: node features are first aggregated to hyperedges to update hyperedge embeddings, which are aggregated back to nodes to update node embeddings. These message-passing-based models have several limitations, including oversmoothing~\cite{chen2022preventing}, high inference latency~\cite{antelmisurvey,zhang2022graphless}, and sensitivity to structural perturbations~\cite{sun2022adversarial, hu2023hyperattack}. For this challenge, we propose Hypergraph-MLP for processing hypergraph-structured data without message passing.%\Note{sc: this sentence is too long}

\textbf{Multilayer perceptions for structural data.} In the literature, many recent works aim to create more efficient and robust models for processing data on graphs using MLPs. These models incorporate the graph structure as part of the supervision during training, either through the GNN-MLP knowledge distillation loss~\cite{zhang2022graphless} or the neighbour contrastive loss~\cite{hu2021graph}. These studies improve MLPs by explicitly considering pairwise node relations, which makes them not well-suited for hypergraph-structured data with higher-order node relations. To our knowledge, our work is the first to demonstrate the effectiveness of MLPs in processing hypergraph-structured data.

\section{Preliminaries}

\textbf{Hypergraphs.} We denote a hypergraph as a triplet $\hhH = \{\V, \E, \hH\}$, where $\V = \{v_1, v_2, \cdots, v_n\}$ is the node set with $|\V|=n$, $\E=\{e_1, e_2, \cdots, e_m\}$ is the hyperedge set with $|\E| = m$, and $\hH \in \{0, 1\}^{n\times m}$ is an incidence matrix in which $\hH_{ij} = 1$ indicates that hyperedge $j$ contains node $i$ and $\hH_{ij} = 0$ otherwise. 

\textbf{Problem formulation.} Let $\X_\V = [\x_{v_1}, \x_{v_2}, \cdots, \x_{v_n}]^{T} \!\in\!\R^{n\times d}$ denote a matrix that contains $d$-dimensional features of nodes on a hypergraph structure $\hH\!\in\!\{0, 1\}^{n\times m}$, and $\Y_\V = [\y_{v_1}, \y_{v_2}, \cdots, \y_{v_n}]^{T}\!\in\!\R^{n\times c}$ be the associated ground-truth label matrix, where $\y_{v_i}\in\{0,1\}^{c}$ is a one-hot encoded label vector. Following previous works~\cite{chien2022you,feng2019hypergraph,bai2021hypergraph}, we focus on the hypergraph node classification task where labels for a set of nodes $\V^{\prime}$ are given at training. We aim to classify nodes in $\V\setminus\V^{\prime}$ during inference.

\section{Learning Framework: Hypergraph-MLP}
In this section, we start by introducing the architecture of the MLP-based model used in our Hypergraph-MLP. Then, we elaborate on how to use the hypergraph smoothness prior to derive the loss function enabling the MLP-based model to use the structural information from $\hH$ without requiring it during inference. Furthermore, we summarize the training and inference process of the proposed Hypergraph-MLP. Finally, we discuss the advantages of Hypergraph-MLP over message-passing-based hypergraph neural networks. To make the following discussion accessible, we denote the node embeddings in the layer $l$ of the MLP-based model as $\Z^{(l)}_\V = [\z_{v_1}^{(l)}, \z_{v_2}^{(l)}, \cdots, \z_{v_n}^{(l)}]^{T} \in \R^{n\times d^{(l)}}$. Following previous papers~\cite{chien2022you,NEURIPS2019_1efa39bc,feng2019hypergraph, bai2021hypergraph}, we initialize $\Z_{\V}^{(0)}$ with $\X_{\V}$. Additionally, we denote the output node embeddings and the associated hyperedge embeddings as $\Z_{\V} = [\z_{v_1}, \z_{v_2}, \cdots, \z_{v_n}]^{T}\in \R^{n\times d^o}$ and $\Z_{\E} = [\z_{e_1}, \z_{e_2}, \cdots, \z_{e_m}]^{T}\in \R^{m\times d^o}$, respectively.

\subsection{MLP-based Model}
\label{sec:mlp}
Inspired by previous works in graph machine learning~\cite{zhang2022graphless,hu2021graph}, we design each layer of the MLP-based model as:
\begin{equation}
\setlength{\abovedisplayskip}{.1pt}
\setlength{\belowdisplayskip}{.1pt}
\label{eq:mlp}
 \Z_{\V}^{(l)} = D(LN(\sigma(\Z_{\V}^{(l-1)}\ttt^{(l)}))),
\end{equation} 
where $D(\cdot)$ is the dropout function, $LN(\cdot)$ is the layer normalization function, $\sigma(\cdot)$ is the activation function, and $\ttt^{(l)}\in\R^{d^{(l-1)}\times d^{(l)}}$ denotes the learnable parameters for layer $l$. For the node classification task, we generate the node label logits by:
\begin{equation}
\setlength{\abovedisplayskip}{.1pt}
\setlength{\belowdisplayskip}{.1pt}
\label{eq:cls}
 \hat{\Y}_{\V} = softmax(\Z_{\V}\W),
\end{equation} 
where $\hat{\Y}_{\V}\in(0,1)^{n\times c}$ denotes the node label logits, $\W\in\R^{d^o\times c}$ denotes some learnable parameters, and $softmax(\cdot)$ is the softmax function. Notably, this MLP-based model only takes the $\X_{\V}$ as input to perform inference. In the next section, we derive a hypergraph-smoothness-based loss function to allow the MLP-based model to integrate the structural information from $\hH$ at training.

\subsection{Hypergraph-Smoothness-based Loss Function}
\label{sec:sth_loss}
In this section, we first model the joint distribution of $\Z_{\V}$ and $\Z_{\E}$ with the given $\hH$ based on a smoothness prior inspired by~\cite{tang2023hypergraph}. We then derive the hypergraph-smoothness-based loss function through maximum likelihood estimation on this joint distribution. 

\SetKwInput{KwLet}{Initialisation}
\begin{algorithm}[t]
\SetAlgoLined
\caption{Hypergraph-MLP For Node Classification}
\textbf{/*Training*/}\\
\KwIn{$\X_{\V}$, $\Y_{\V}$, $\hH$, $\V^{\prime}$, the number of layers $L$, the number of the training iterations $T$.}
\KwOut{The optimal model parameters $[\ttt^{(1)^\star}, \cdots, \ttt^{(L)^\star}, \W^{\star}]$.}
\KwLet{Randomly initialise the model parameters $[\ttt^{(1)}, \cdots, \ttt^{(L)}, \W]$, and set $\Z_{\V}^{(0)}$ as $\X_{\V}$.}
\For{$t=1:T$ }
{
\For{$l=1:L$ }
{
Update the node embeddings by Eq.~(\ref{eq:mlp}).\\
}
Generate the node label logits $\hat{\Y}_{\V}$ by Eq.~(\ref{eq:cls}).\\
Update $[\ttt^{(1)}, \cdots, \ttt^{(L)}, \W]$ by minimizing Eq.~(\ref{eq:overall_loss}) with the gradient descent.
}
\vspace{1mm}
\textbf{/*Inference*/}\\
\KwIn{$\X_{\V}$ and $[\ttt^{(1)^\star}, \cdots, \ttt^{(L)^\star}, \W^{\star}]$.}
\KwOut{Predicted node label logits $\hat{\Y}_{\V}$.}
\For{$l=1:L$ }
{
Update the node embeddings by Eq.~(\ref{eq:mlp}).\\
}
Generate the label logits $\hat{\Y}_{\V}$ by Eq.~(\ref{eq:cls}).
\label{al:hypergraph_mlp}
\vskip -0.02in
\end{algorithm}

\textbf{Hypergraph node embedding smoothness prior.} Taking the inspiration from a recent work~\cite{tang2023hypergraph}, we propose a smoothness prior on the hypergraph node embedding: \textit{embeddings of nodes in a hyperedge are highly correlated via the hyperedge connecting these nodes.} Under this prior, we capture the relation between nodes and hyperedges in the latent space by an incidence graph~\cite{godsil2001algebraic}. For the hypergraph $\hhH=\{\V,\E,\hH\}$, we construct the unique incidence graph as a bipartite graph $\G=\{\V\bigcup\V', \E_{\G}, \LL_{\hH}\}$. Here $v_i\in\V$ is node $v_i$ in $\hhH$, $v_{e_{j}}\in\V'$ corresponds to the hyperedge $e_j$ in $\hhH$. An edge exists between $v_i$ and $v_{e_{j}}$ if and only if $e_{j}$ contains $v_i$ in $\hhH$. The structure of $\G$ is represented as a graph Laplacian matrix:
\begin{equation}
\setlength{\abovedisplayskip}{.1pt}
\setlength{\belowdisplayskip}{.1pt}
\begin{small}
\LL_{\hH} = \begin{bmatrix}
\diag(\hH\textbf{1}_{m}) & -\hH \\
-\hH^T & \diag(\hH^{T}\textbf{1}_{n})
\end{bmatrix},
\end{small}
\label{eq:laplacian}
\end{equation}
where $\LL_{\hH}\in\R^{(n+m)\times(n+m)}$, $\diag(\cdot)$ represents the transformation of a vector into a diagonal matrix, and $\textbf{1}_{n}\in\{1\}^{n}$ and $\textbf{1}_{m}\in\{1\}^{m}$ are two all-one vectors. Following the literature of graph signal processing~\cite{kalofolias2017large,dong2016learning,pu2021learning}, with the incidence graph formulated as Eq.~(\ref{eq:laplacian}) the underlying distribution of $\Z_\V$ and $\Z_\E$ can be modelled as:
\begin{equation}
\setlength{\abovedisplayskip}{.1pt}
\setlength{\belowdisplayskip}{.1pt}
\begin{small}
[\Z_{\V}^{T}, \Z_{\E}^{T}]^{T}\sim \N(\textbf{0}, \LL_{\hH}^{\dagger}),
\end{small}
\label{eq:gaussian}
\end{equation}
where $\LL_{\hH}^{\dagger}$ is the pseudoinverse of $\LL_{\hH}$.

\textbf{Loss function via maximum likelihood estimation.} Our goal is to derive a loss function from the likelihood of $[\Z_{\V}^{T}, \Z_{\E}^{T}]^{T}$, thus allowing the MLP to utilize the hypergraph structural information by generating node embeddings that follow our smoothness prior. With Eq.~(\ref{eq:gaussian}) and setting $\hH$ to obey a uniform distribution, we have $ p(\hH|\Z_{\V},\Z_{\E})\!\propto\!e^{ -\frac{1}{2}[\Z_{\V}^{T}, \Z_{\E}^{T}]\LL_{\hH}[\Z_{\V}^{T}, \Z_{\E}^{T}]^{T}}$. Omitting the constant term, the negative log likelihood function of $[\Z_{\V}^{T}, \Z_{\E}^{T}]^{T}$ is:
\begin{eqnarray}
\setlength{\abovedisplayskip}{.1pt}
\setlength{\belowdisplayskip}{.1pt}
{\cal L}(\Z_{\V},\Z_{\E}|\hH) & \propto & [\Z_{\V}^{T},\Z_{\E}^{T}]\LL_{\hH}[\Z_{\V}^{T}, \Z_{\E}^{T}]^{T}  \nonumber\\
& = & \sum_{i=1}^{m}\sum_{v_j\in e_i}||\z_{e_i}-\z_{v_j}||^{2}_2.
\label{eq:inter1}
\end{eqnarray}
To compute Eq.~(\ref{eq:inter1}), we need $\Z_{\E}$, which could be generated using $\X_{\V}$ and $\hH$ as input. However, as discussed in Section~\ref{sec:mlp}, the MLP-based model only takes $\X_{\V}$ as input. Hence, inspired by~\cite{tang2023hypergraph}, we derive a lower bound for Eq.~(\ref{eq:inter1}) based on the triangle inequality~\cite{abramowitz1988handbook}:
\begin{eqnarray*}
\setlength{\abovedisplayskip}{.1pt}
\setlength{\belowdisplayskip}{.1pt}
\sum_{i=1}^{m}\sum_{v_j\in e_i}||\z_{e_i}-\z_{v_j}||^{2}_2 & \geq  & \sum_{i=1}^{m}(||\z_{e_i}-\z_{v_{a_i}}||^{2}_2+||\z_{e_i}-\z_{v_{b_i}}||^{2}_2)\nonumber\\
%& \geq& \sum_{i=1}^{m}||\z_{v_{a_i}} - \z_{v_{b_i}}||_{2}\nonumber\\ 
& \geq&\sum_{i=1}^{m}\mathop{\max}\limits_{v_j, v_k \in e_i}(||\z_{v_j} - \z_{v_k}||_{2}^{2}),
\end{eqnarray*}
where $v_{a_i}$ and $v_{b_i}$ are the two most distant nodes within $e_i$ in the output feature latent space. With the lower bound shown above, we formulate the hypergraph-smoothness-based loss function as:
\begin{equation}
\setlength{\abovedisplayskip}{.1pt}
\setlength{\belowdisplayskip}{.1pt}
\label{eq:sth_loss}
\ell_{smooth} = \frac{1}{m}\sum_{i=1}^{m}\mathop{\max}\limits_{v_j, v_k \in e_i}(||\z_{v_j} - \z_{v_k}||_{2}^{2}).
\end{equation} 
Minimizing this loss function allows the model in Eq.~(\ref{eq:mlp}) to generate node embeddings that match the distribution in Eq.~(\ref{eq:gaussian}), thereby integrating the hypergraph structural information.

\subsection{Training \& Inference}
For node classification, we train the MLP-based model using a combination of the aforementioned hypergraph-smoothness-based loss function and the cross-entropy loss function for node classification:
\begin{equation}
\setlength{\abovedisplayskip}{.1pt}
\setlength{\belowdisplayskip}{.1pt}
\label{eq:ce_loss}
\ell_{CE} = -\frac{1}{|\V^{\prime}|}\sum_{v_i\in\V^{\prime}}\y_{v_i}^{T}\log(\hat{\y}_{v_i}),
\end{equation} 
where $\V^{\prime}$ is a set containing selected training nodes, and $|\V^{\prime}|$ is the number of the selected training nodes. The overall loss function is:
\begin{equation}
\setlength{\abovedisplayskip}{2pt}
\setlength{\belowdisplayskip}{2pt}
\label{eq:overall_loss}
\ell_{overall} = \ell_{CE} + \alpha\ell_{smooth}.
\end{equation}
where $\alpha$ is a coefficient to balance the two losses. In practice, our model is trained by using gradient descent to minimize the Eq.~(\ref{eq:overall_loss}). The proposed framework is summarized in Algorithm~\ref{al:hypergraph_mlp}. 

\subsection{Discussion}
\textbf{Oversmoothing.} A recent study~\cite{chen2022preventing} highlights a common challenge encountered in the message-passing-based hypergraph neural networks: the oversmoothing issue. In the literature~\cite{chen2022preventing, rusch2023survey}, the oversmoothing issue in the MPNN is attributed to node feature aggregation within each forward-propagation layer. This operation tends to promote the similarity of embeddings among connected nodes, leading to increasingly homogeneous node embeddings as the depth of the MPNN increases. Notably, in contrast to MPNNs, our MLP-based model, as delineated in Eq.~(\ref{eq:mlp}), does not employ node feature aggregation within its forward-propagation layers. Consequently, a deeper MLP-based model does not necessarily lead to greater similarity among node embeddings. This crucial distinction renders the Hypergraph-MLP inherently immune to the oversmoothing issue.

\textbf{Inference speed.} The application of the message-passing-based hypergraph neural networks to real-world scenarios faces challenges due to high inference latency~\cite{antelmisurvey,zhang2022graphless, hu2021graph}. Let $n$ be the number of nodes, $m$ be the number of hyperedges, and $L$ be the number of layers. The computational complexity of a hypergraph neural network is $\mathcal{O}(Ln + Lm)$, as it involves feature aggregation for every node and hyperedge in each layer. In contrast, the Hypergraph-MLP performs inference solely via feed-forward propagation, as formulated in Eq.~(\ref{eq:mlp}). Consequently, its computational complexity is $\mathcal{O}(Ln)$, which is significantly lower especially when dealing with datasets rich in hyperedges, such as DBLP as demonstrated in Table~\ref{tab:pro_data}. In the next section, we empirically illustrate that the reduction at inference complexity facilitates the Hypergraph-MLP in achieving lower inference latency compared to existing hypergraph neural networks.

\textbf{Inference robustness.} The significant dependence on the hypergraph structure for message passing renders current hypergraph neural networks vulnerable to structural perturbations at inference. For instance, the introduction of fake hyperedges during inference can lead well-trained hypergraph neural networks to generate baffling results~\cite{sun2022adversarial,hu2023hyperattack}. In contrast, Hypergraph-MLP implicitly takes into account the hypergraph structure, thus removing its dependence on the structure during inference. In the next section, we present empirical evidence to demonstrate that this property enhances the robustness of Hypergraph-MLP compared to existing hypergraph neural networks in the presence of structural perturbations at inference.

\begin{table}[t]
\begin{center}
\caption{Properties of datasets.}
\vskip -0.1in
\label{tab:pro_data}\resizebox{.9\columnwidth}{!}{
\begin{tabular}{ccccccccc}
\hline
 &Cora & Citeseer & Pubmed & DBLP & 20News & NTU2012 & House
\\
\hline
$|\V|$ & 2708&  3312&  19717& 41302&  16242&  2012& 1290
\\
$|\E|$ & 1579&  1079&  7963& 22363&  100&  2012& 341
\\
\# features & 1433&  3703&  500& 1425&  100&  100& 100
\\
\# class & 7&  6&  3& 6&  4&  67& 2
\\
Homophily & 0.84 &  0.78 & 0.79 & 0.88 & 0.49 & 0.81 & 0.52 
\\
\hline
\end{tabular}}
\end{center}
\vskip -0.3in
\end{table}

\begin{table*}[h]
\begin{center}
\caption{Comparison with baselines on clean datasets. Mean testing \textbf{ACC} (\%) ± standard deviation from 20 runs.}
\vskip -0.1in
\label{tab:real-world_a}\resizebox{0.9\columnwidth}{!}{
\begin{tabular}{c|ccccccc|c}
\hline
&Cora & Citeseer & Pubmed & DBLP & 20News & NTU2012 & House & Avg Mean
\\
\hline
HyperGCN & 78.45 ± 1.26 &
  71.28 ± 0.82 &
  82.84 ± 8.67 &
  89.38 ± 0.25 &
  81.05 ± 0.59 &
  56.36 ± 4.86 &
  78.22 ± 2.46 & 
76.80
\\
HGNN & 79.39 ± 1.36 &
  72.45 ± 1.16 &
  86.44 ± 0.44 &
  91.03 ± 0.20 &
  80.33 ± 0.42 &
  87.72 ± 1.35 &
  66.16 ± 1.80 & 80.50
\\
HCHA & 79.14 ± 1.02 &
  72.42 ± 1.42 &
  86.41 ± 0.36 &
  90.92 ± 0.22 &
  80.33 ± 0.80 &
  87.48 ± 1.87 &
  67.91 ± 2.26 & 80.66
\\
UniGCNII & 78.81 ± 1.05 &
73.05 ± 2.21 &
  88.25 ± 0.40 &
  \textbf{91.69 ± 0.19} &
  81.12 ± 0.67 &
  \textbf{89.30 ± 1.33} &
  80.65 ± 1.96 & 83.27
\\
AllDeepSets & 76.88 ± 1.80 &
  70.83 ± 1.63 &
  \textbf{88.75 ± 0.33} &
  91.27 ± 0.27 &
  81.06 ± 0.54 &
  88.09 ± 1.52 &
  80.70 ± 1.59 & 82.51
\\
AllSetTransformer & 78.59 ± 1.47 &
  73.08 ± 1.20 &
  88.72 ± 0.37 &
  91.53 ± 0.23 & 
  81.38 ± 0.58 &
  88.69 ± 1.24 &
  83.14 ± 1.92 & 83.59
\\
MLP & 74.99 ± 1.49&  72.31 ± 1.28&   87.69 ± 0.59& 85.53 ± 0.27&  81.70 ± 0.49&  87.89 ± 1.36& 83.78 ± 1.96 & 81.98
\\
\hline
Hypergraph-MLP &\textbf{79.80 ± 1.82} &  \textbf{73.90 ± 1.57}& 87.89 ± 0.55& 90.29 ± 0.26&  \textbf{81.75 ± 0.41}&  88.42 ± 1.32&\textbf{84.03 ± 1.75} & \textbf{83.72}
\\
\hline
\end{tabular}}
\end{center}
\vskip -0.2in
\end{table*}

\begin{table*}[h]
\begin{center}
\caption{Comparison with baselines on clean datasets. Mean \textbf{inference time} (ms) ± standard deviation from 60000 runs.}
\vskip -0.1in
\label{tab:real-world_t}\resizebox{.9\columnwidth}{!}{
\begin{tabular}{c|ccccccc|c}
\hline
&Cora & Citeseer & Pubmed & DBLP & 20News & NTU2012 & House & Avg Mean
\\
\hline
HyperGCN & 0.46 ±
0.10
  &  0.49 ±
0.10& 
0.60 ± 0.07 &
  1.19 ± 0.21
 & 
0.82 ± 0.30 &
  0.47 ± 0.10 &
  0.50 ± 0.10 & 0.65 
\\
HGNN & 
1.40 ± 0.11&
1.41 ± 0.33&
3.33 ± 0.26 &
4.36 ± 0.41& 
1.39 ± 0.33&
1.26 ± 0.11&
1.43 ± 0.14 & 2.08
\\
HCHA & 1.33 ±
0.38 &1.37 ±
0.14&
  3.43 ±
0.17&
  8.08 ±
0.76& 
1.45 ±
0.14 &
  1.29 ±
0.15
 &1.46 ±
0.16 &2.63
\\
UniGCNII & 3.01 ± 0.14 &
 0.71 ± 0.10 &
 0.85 ± 0.07   &
 21.15 ± 0.64 & 
  3.79 ± 0.18
 &
  1.48 ± 0.08
  &
 0.77 ± 0.13 & 4.54
\\
AllDeepSets & 2.02 ±
0.15 &
  2.40 ±
0.14 &
  8.90 ±
1.09 &
  21.51 ±
0.43 & 
 7.68 ±
0.57
 &
  1.96 ±
0.09 &
  1.39 ±
0.13 & 6.55

\\
AllSetTransformer & 1.91 ±
1.34 &
  3.01 ±
1.32 &
  4.75 ±
1.33 &
  21.21 ±
1.42 & 
  4.30 ±
1.57
 &
  1.90 ±
1.32 &
  2.14 ±
1.29 & 5.60
\\
\hline
Hypergraph-MLP &\textbf{0.45 ± 0.11} &  \textbf{0.21 ±
0.07}& 
\textbf{0.19 ± 0.04}& \textbf{0.47 ± 0.17}&  \textbf{0.28 ±
0.05}&  \textbf{
0.36 ±
0.08}&\textbf{
0.28 ± 0.05} & \textbf{0.32}
\\
\hline
\end{tabular}}
\end{center}
\vskip -0.3in
\end{table*}
\section{Experiments}

\subsection{Experimental Settings}

\textbf{Datasets.} We use seven public datasets, including academic hypergraphs (Cora, Citeseer, Pubmed, and DBLP), adapted from~\cite{NEURIPS2019_1efa39bc}, 20News from UCI's Categorical Machine Learning Repository~\cite{dua2017uci}, NTU2012 from computer vision~\cite{chen2003visual}, and House from politics~\cite{chodrow2021hypergraph}. For the House dataset, lacking node features, we follow~\cite{chien2022you} and use Gaussian random vectors instead, where the standard deviation of the added Gaussian features is set as $0.6$. Notably, the datasets 20News and House represent examples of heterophilic hypergraphs, while Cora, Citeseer, Pubmed, DBLP, and NTU2012 exemplify homophilic hypergraphs. Details of the datasets used are in Table~\ref{tab:pro_data}.  

\textbf{Baselines.} We compare Hypergraph-MLP with six message-passing hypergraph neural networks: HyperGCN~\cite{NEURIPS2019_1efa39bc}, HGNN~\cite{feng2019hypergraph}, HCHA~\cite{bai2021hypergraph}, UniGCNI~\cite{ijcai21-UniGNN}, AllDeepSets~\cite{chien2022you}, and AllsetTransformer~\cite{chien2022you}; and a standard MLP, differing from our Hypergraph-MLP only in the loss function, using Eq.~(\ref{eq:ce_loss}) instead of Eq.~(\ref{eq:overall_loss}).

\textbf{Metric.} In accordance with prior research~\cite{chien2022you,NEURIPS2019_1efa39bc,feng2019hypergraph,bai2021hypergraph,ijcai21-UniGNN}, we evaluate the performances of baselines and our method by accuracy (ACC: $\%$), which is defined as the ratio between the number of correct predictions to the total number of predictions.

\textbf{Our implementation.} Following a previous work~\cite{chien2022you}, we randomly partitioned the dataset into training, validation, and test sets using a 50\%/25\%/25\% split. All experiments were performed on an RTX 3090 using PyTorch. We identified the optimal value of $\alpha$ in Eq.~(\ref{eq:overall_loss}) through grid search. Our code is available at: https://github.com/tbh-98/Hypergraph-MLP.

\subsection{Results \& Analysis}
\begin{figure}[t!]
\begin{center}
\centering
\subfigure[Cora.]{
         	\includegraphics[width=.4\columnwidth]{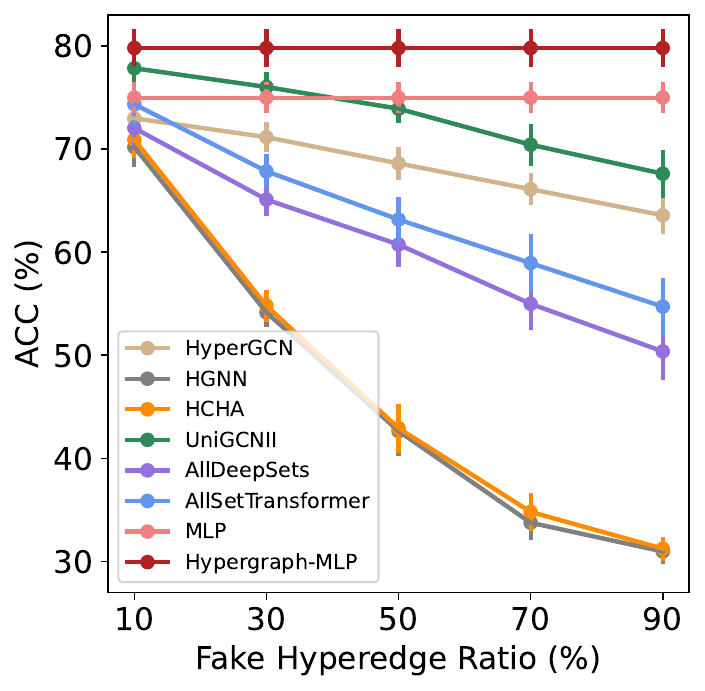}
 	}
\subfigure[DBLP.]{
        	\includegraphics[width=.4\columnwidth]{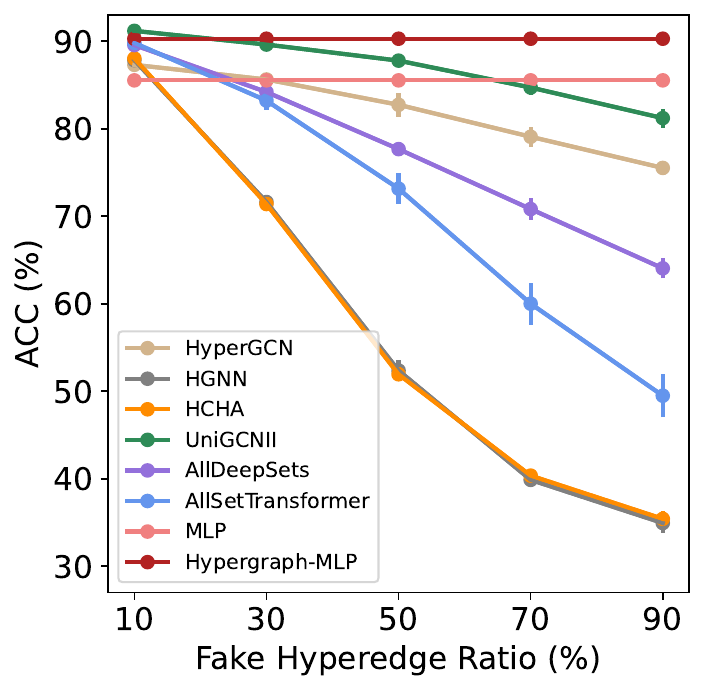}
 	}
\vspace{-3mm}
\caption{Comparision with baselines on the perturbed Cora and DBLP. Mean \textbf{ACC} (\%) ± standard deviation from 20 runs.}
\label{fig:acc}
\end{center}
\vspace{-5mm}
\end{figure}

\textbf{Performance comparison on clean datasets.} The results for real-world hypergraph node classification benchmarks, conducted without considering structural perturbations, are summarized in Table~\ref{tab:real-world_a} and Table~\ref{tab:real-world_t}. In Table~\ref{tab:real-world_a}, we present the ACC comparison of the proposed Hypergraph-MLP against all baseline methods. Our results demonstrate that Hypergraph-MLP not only outperforms the baseline methods across four datasets (Cora, Citeseer, 20News, and House) but also achieves the highest average mean ACC over all of the datasets. Furthermore, Hypergraph-MLP consistently outperforms the standard MLP across all the datasets. These findings underscore the efficacy of the hypergraph-smoothness-based loss function in effectively enabling the MLP-based model to leverage hypergraph structural information, all while not explicitly using the structure itself as input for message passing. As illustrated in Table~\ref{tab:real-world_t}, Hypergraph-MLP demonstrates the fastest inference speed on all the seven chosen datasets compared to existing message-passing-based hypergraph neural networks. Notably, across the seven datasets, Hypergraph-MLP achieves an average mean inference time that is only $49\%$ of the time required by the fastest message-passing-based model (HyperGCN) and a mere $5\%$ of the time needed by the most time-consuming message-passing-based model (AllDeepSets). This attribute renders Hypergraph-MLP well-suited for applications with stringent latency constraints, such as autonomous driving and recommender systems, in contrast to previous hypergraph neural networks.

\textbf{Performance comparison on perturbed datasets.} We present results on perturbed Cora and DBLP in Fig.~\ref{fig:acc}. We perturb hypergraph structures by replacing original hyperedges with fake ones during inference. Moreover, we define the fake hyperedge ratio as $\frac{n_{f}}{n_{o}} * 100\%$, where $n_{f}$ represents the number of fake hyperedges in the perturbed structure, and $n_{o}$ is the number of hyperedges in the original structure. The results are consistent across perturbed Cora and DBLP: compared with message-passing-based hypergraph neural networks, the Hypergraph-MLP consistently demonstrates better robustness against structural perturbations. Eliminating the reliance on hypergraph structure during inference makes Hypergraph-MLP unaffected by structural perturbations introduced exclusively at inference. This characteristic positions Hypergraph-MLP as a robust choice for applications vulnerable to structure evasion attacks.

\section{Conclusion}
We propose a novel hypergraph learning framework, Hypergraph-MLP. To our knowledge, it is the first framework to efficiently incorporate the hypergraph structural information without using the structure itself as input to perform message passing. The key of Hypergraph-MLP is a hypergraph-smoothness-based loss function enabling an MLP-based model to learn the node embedding distribution without explicit reliance on hypergraph structures during forward propagation. Compared to existing hypergraph neural networks, the advantages of Hypergraph-MLP are threefold. Firstly, it helps avoid oversmoothing. Secondly, it exhibits superior computational efficiency during inference. Finally, it demonstrates better robustness against structural perturbations at inference. Our framework can potentially benefit the hypergraph research community by enhancing the efficiency and robustness of learning on hypergraphs.

\textbf{Limitation and future work.} 
The Hypergraph-MLP is currently limited to transductive learning, with the hyperparameter $\alpha$ in Eq.~(\ref{eq:overall_loss}) can only be optimized through grid search. The potential future works include extending to the inductive learning setting and improving the hyperparameter selection mechanism.

\section*{Acknowledgment}
This research is supported by NSFC under Grant 62171276 and 62211530109, as well as the Science and Technology Commission of Shanghai Municipal under Grant 21511100900, 22511106101, and 22DZ2229005. X.D. acknowledges support from the Oxford-Man Institute of Quantitative Finance, the EPSRC (EP/T023333/1), and the Royal Society (IEC\textbackslash NSFC\textbackslash 211188).

\bibliographystyle{IEEEbib}
\bibliography{refs}

\end{document}